\documentclass[letterpaper, 10pt, conference]{ieeeconf}  

\IEEEoverridecommandlockouts                              
\usepackage[utf8]{inputenc}
\usepackage[T1]{fontenc}
\usepackage{graphicx}
\usepackage{listings}
\usepackage{breqn}
\usepackage{booktabs}
\usepackage{pythontex}
\usepackage{comment}
\graphicspath{ {./} }

\title{\LARGE \bf
Improving Reliable Navigation under Uncertainty via Predictions Informed by Non-Local Information
}

\author{Raihan Islam Arnob and Gregory J. Stein\thanks{R. Arnob and G. Stein are with the Department of Computer Science, George Mason University, USA, \{\texttt{rarnob}, \texttt{gjstein}\}\texttt{@gmu.edu}}}

\begin{document}
\maketitle
\thispagestyle{empty}
\pagestyle{empty}

\begin{abstract}
We improve reliable, long-horizon, goal-directed navigation in partially-mapped environments by using non-locally available information to predict the goodness of temporally-extended actions that enter unseen space.
Making predictions about where to navigate in general requires non-local information: any observations the robot has seen so far may provide information about the goodness of a particular direction of travel.
Building on recent work in learning-augmented model-based planning under uncertainty, we present an approach that can both rely on non-local information to make predictions (via a graph neural network) and is reliable by design: it will always reach its goal, even when learning does not provide accurate predictions. 
We conduct experiments in three simulated environments in which non-local information is needed to perform well.
In our large scale university building environment, generated from real-world floorplans to the scale, we demonstrate a 9.3\% reduction in cost-to-go compared to a non-learned baseline and a 14.9\% reduction compared to a learning-informed planner that can only use local information to inform its predictions.
\end{abstract}
\section{Introduction}
We focus on the task of goal-directed navigation in a partially-mapped environment, in which a robot is expected to reach an unseen goal in minimum expected time.
Often modeled as a Partially Observable Markov Decision Process (POMDP)~\cite{kaelbling1998}, long-horizon navigation under uncertainty is computationally demanding, and so many strategies turn to learning to make predictions about unseen space and thereby inform good behavior.
To perform well, a robot must understand how parts of the environment the robot cannot currently see (i.e., non-locally available information) inform where it should go next, a challenging problem for many existing planning strategies that rely on learning.

\begin{figure}[t]
\vspace{6pt}
  \includegraphics[width=.48\textwidth]{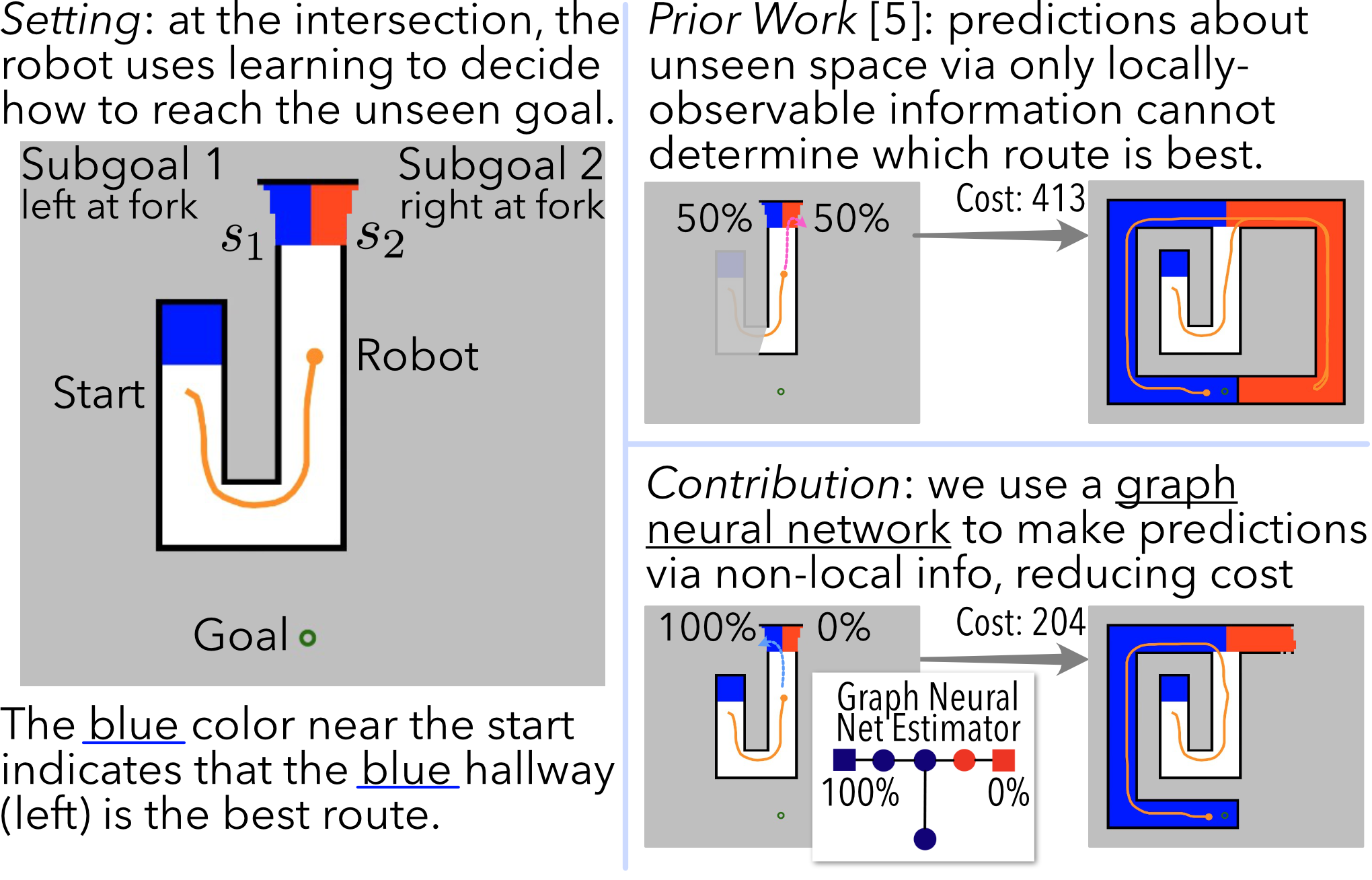}
  \caption{
  \textbf{Overview: non-local information is often essential for good navigation in a partial map.} Our LSP-GNN approach uses a graph neural network to make predictions about unseen space via both local and non-local information and integrates these into the Learning over Subgoals model-based planning abstraction~\cite{pmlr-v87-stein18a,bradley2021} to improve reliable navigation.
  }
  \label{fig:intro-explain}
\end{figure}

Consider the simple scenario from our \emph{J-Intersection} environment shown in Fig.~\ref{fig:intro-explain}: information at the center of the map (the color of that region) informs whether the robot should travel left or right; optimal behavior involves following the hallway whose color matches that of the center of the map.
As this color is not visible from the intersection, a robot must remember what the space looked like around the corner to perform well and learn how that information relates to its decision.
More generally, many real-world environments require such understanding, a particularly challenging task for building-scale environments. In this work, we aim to allow a robot to retain non-local knowledge and learn to use it to make predictions that inform where it should travel next.



Recently, learning-driven approaches---including many model-free approaches trained via deep reinforcement learning~\cite{MERLIN2018_Greg_Wayne, mirowski2018}---have demonstrated the capacity to perform well in this domain.
However, in the absence of an explicit map for the robot to use to keep track of where it has yet to go, many such approaches are unreliable, lacking guarantees that they will reach the goal~\cite{pfeiffer2016}.
Moreover, these approaches struggle to reason far enough into the future to understand the impact of their actions and thus perform poorly and can be brittle and unreliable for long-horizon planning.


The recent Learning over Subgoals planning approach (LSP)~\cite{pmlr-v87-stein18a} introduces a high-level abstraction for planning in a partial map that allows for both state-of-the-art performance and reliability-by-design.
In LSP, actions correspond to exploration of a particular region of unseen space.
Learning (via a fully-connected neural network) is used to estimate the goodness of exploratory actions, including the likelihood an exploration will reveal the unseen goal.
These predictions inform model-based planning and are thus used to compute expected cost.
LSP overcomes two problems: (1) its state and action abstraction allows for learning-informed reasoning far into the future and (2) it is guaranteed to reach the goal if there exists a viable path.
However, LSP is limited: its ability to make predictions about unseen space only makes use of locally observable information, limiting its performance.

In this paper, we extend the Learning over Subgoals Planner (LSP-Local), replacing its learning backend with a Graph Neural Network (LSP-GNN), affording reliable learning-informed planning capable of using both local and non-local information to make predictions about unseen space and thus improve performance in complex navigation scenarios in building-scale environments.
Using a graph representation of the partial map---constructed via a map skeleton~\cite{zhang1984} so as to preserve topological structure---we demonstrate that our GNN allows for accurate predictions of unseen space using non-local information.
Additionally, we demonstrate that our LSP-GNN planner improves performance over the original LSP-Local planner while retaining guarantees on reliability: i.e., the robot always reaches the goal.
We show the effectiveness of our approach in our simulated \emph{J-Intersection}, \emph{Parallel Hallway}, and \emph{University Building} environments, in the latter yielding improvements of 9.3\% and 14.9\% (respectively) over non-learned and learned baselines. 

\section{Related Works}\label{sec:related-works}

\textbf{Planning under Uncertainty}
POMDPs~\cite{kaelbling1998, littman1997, thrun2005,parascandolo-DCMCTS} have been used to represent navigation and exploration tasks under uncertainty, yet direct solution of the model implicit in the POMDP is often computationally infeasible.
To mitigate this limitation, many approaches to planning rely on learning to inform behavior~\cite{pfeiffer2016, richter2014,ross2012}, yet only plan a few time steps into the future and so are not well-suited to long-horizon planning problems.
Some reinforcement learning approaches that deal with partially observed environments~\cite{DuanSCBSA16, yang2021,gupta2019cognitive, zhang2017deep, TaiPL17, MirowskiPVSBBDG16} are also limited to fairly small-scale environments.
The MERLIN agent~\cite{MERLIN2018_Greg_Wayne} uses a differentiable neural computer to recall information over much longer time horizons than is typically possible for end-to-end-trained model-free deep reinforcement learning systems.
However, the reinforcement learning approaches~\cite{MERLIN2018_Greg_Wayne,Kober2014, henderson2017} can be difficult to train and lacks plan completeness, making it somewhat brittle in practice.
Our proposed work improves long-horizon planning under uncertainty learning the relational properties from the non-local observation of the environment with the guarantee of completeness.

\textbf{Graph Neural Networks and Planning}
Battaglia et al.~\cite{peter2018} present a survey of GNN approaches, demonstrating how GNNs can be used for relational reasoning and exhibit combinatorial generalization, opening numerous opportunities for learning over structured and relational data.
Zhou et al.~\cite{zhou2018} show how GNNs have been used in the field of modeling physics systems, learning molecular fingerprints, predicting protein interface, classifying diseases, and many others.
GNNs are fast to evaluate on sparse graphs and have shown capacity to generalize effectively in multiple domains \cite{peter2018,duvenaud2015gcnmolecule,monti2017geometric}.
Moreover, GNNs have recently been used to accelerate task and motion planning~\cite{kim2019guidedtamp,kim2021representation} and to inform other problems of interest to robotics: joint mapping and navigation \cite{chen2020gnnexplore}, object search in previously-seen environments~\cite{kurenkov2021semantic}, and modeling physical interaction~\cite{kossen2020structured}.
In particular, Chen et al.~\cite{fanfei2021} propose a framework that uses GNN in conjunction with deep reinforcement learning to address the problem of autonomous exploration under localization uncertainty for a mobile robot with 3D range sensing.
\section{Problem Formulation}
\label{sec:prob-form}
Our robot is tasked to reach an unseen goal in a partially-mapped environment in minimum expected cost (distance).
The synthetic robot is equipped with a semantically-aware planar laser scanner, which it can use to both localize and update its partial semantic-occupancy-grid map of its local surroundings, limited by range and obstacle occlusion.
As the robot navigates the partially-mapped environment, it updates its belief state $b_t$ to include newly-revealed space and its semantic class.

Formally, we represent this problem as a Partially Observable Markov Decision Process~\cite{kaelbling1998,littman1997} (POMDP). The expected cost $Q$ under this model can be written via a belief space variant of the Bellman equation~\cite{Pineau-2002-8519}:
\begin{equation}
\label{eq:POMDP}
\begin{split}
    Q(b_t,a_t) = \sum_{b_{t+1}} P(b_{t+1}|b_t,a_t)\Big[R(b_{t+1},b_t,a_t) \\ + \min_{a_{t+1} \in \mathcal{A}(b_t+1)}Q(b_{t+1},a_{t+1})\Big],
\end{split}
\end{equation}
where $R(b_{t+1},b_t,a_t)$ is the cost of reaching belief state $b_{t+1}$ from $b_t$ by taking action $a_t$ and $P(b_{t+1}|b_t, a_t)$ is the transition probability.
\begin{figure}[t]
  \includegraphics[width=.48\textwidth]{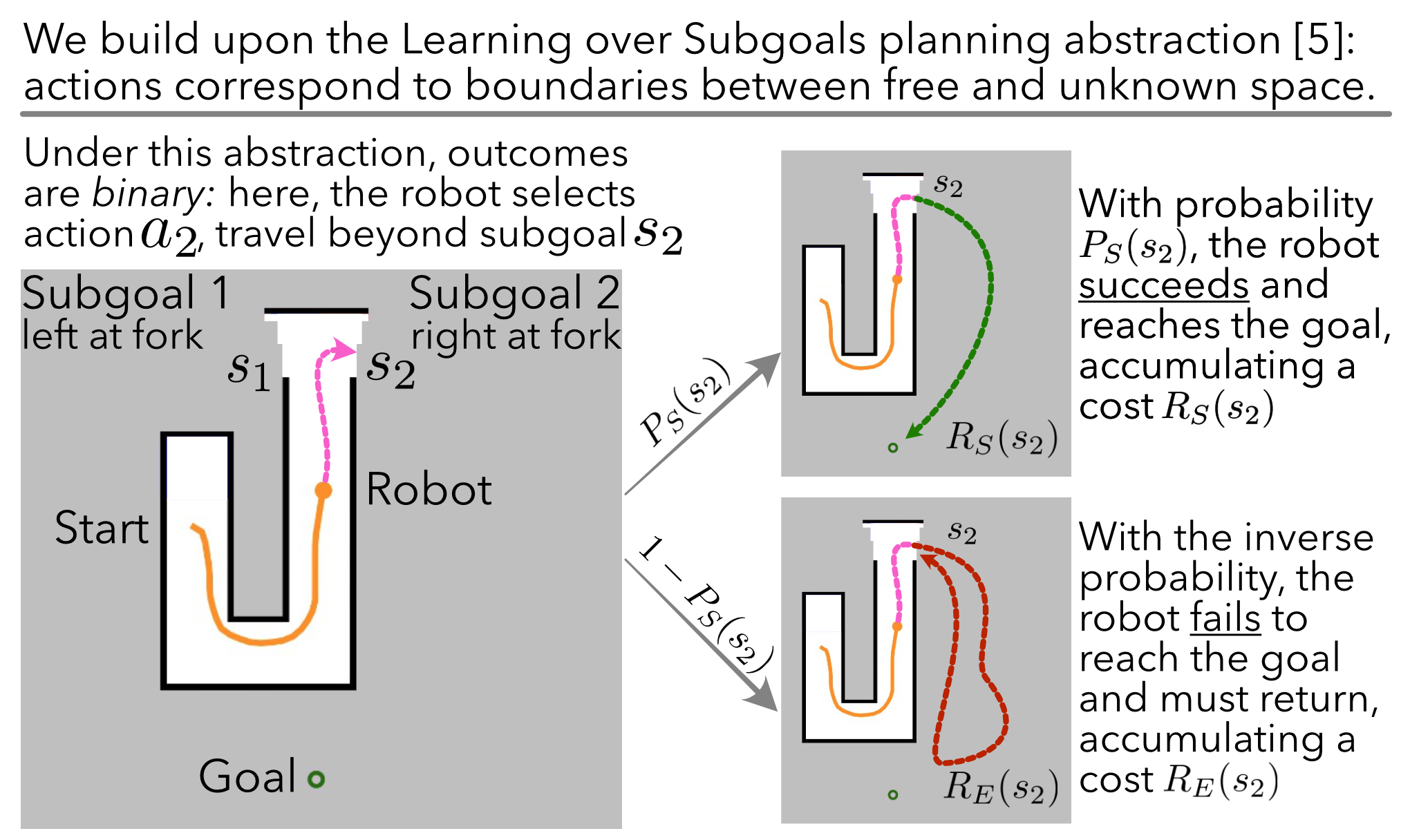}
    \caption{
    \textbf{Our robot's actions correspond to boundaries between free and unseen space.}
    The robot can leave observed space through either boundary: via subgoal $s_1$ or $s_2$. Upon selecting action $a_2$, the robot reaches the goal with probability $P_S$ and incurs an expected cost $R_S$, or is turned back (probability $1-P_S$), accumulates cost $R_E$ and selects another action.
    }
  \label{fig:lsp-example}
\end{figure}

\section{Preliminaries: Model-based Planning under Uncertainty via Learning over Subgoals}\label{sec:lsp}
As Eq.~\eqref{eq:POMDP} cannot be solved directly, our robot instead relies on the recent Learning over Subgoals Planning (LSP) approach~\cite{pmlr-v87-stein18a} to determine the robot's behavior.
LSP introduces a model-based planning abstraction that alleviates the computational requirements of POMDP planning, affording both reliability and good performance informed by predictions about unseen space from learning.

For LSP planning, actions available to the robot correspond to navigation to \emph{subgoals}---each associated with a boundary between free and unknown space---and then exploration beyond in an effort to reach the unseen goal.
Consistent with this action abstraction, planning under the LSP model is done over an abstract belief state: a tuple $b_t = \{m_t, q_t\}$, where $m_t$ is the current map of the environment, and $q_t$ is the robot pose.
Each high-level action $a_t \in \mathcal{A}(\{m_t, q_t\})$ has a binary outcome: with probability $P_S(a_t)$, the robot \emph{succeeds} in reaching the goal or (with the inverse probability $1 - P_S(a_t)$) fails to reach the goal.
Upon selecting an action $a_t$, the robot must first move through known space to the boundary, accumulating a cost $D(m_t, q_t, a_t)$.
If the robot succeeds in reaching the goal, it accumulates a \emph{success cost} $R_S(a_t)$, the expected cost for the robot to reach the goal, and no further navigation is necessary.
Otherwise, the robot accumulates an \emph{exploration cost} $R_E(a_t)$, the expected cost of exploring the region beyond the subgoal of interest and needing to turn back, and must subsequently choose another action $a_{t+1} \in A_{t+1} \equiv \mathcal{A}(\{m_t, q(a_t)\})\setminus \{ a_t \}$.

Under this LSP planning model, the expected cost of taking an action $a_t$ from belief state $b_t = \{ m_t, q_t\}$ is
\begin{equation}\label{eq:lsp-planning}
\begin{split}
    Q(&\{m_t, q_t\}, a_t\in  \mathcal{A}) = D(m_t, q_t, a_t) + P_S(a_t) R_S(a_t) \\
    & + (1-P_S(a_t)) \left[R_E(a_t) + \min_{a_{t+1}}Q(\{m_t, q(a_t)\},a_{t+1}) \right]
    \end{split}
\end{equation}
While the known-space distance $D(m_t, q_t, a_t)$ can be calculated directly from the observed map using A$^{\!*}$ or RRT$^*$, the \emph{subgoal properties} $P_S(a_t)$, $R_S(a_t)$, and $R_E(a_t)$ for each subgoal are estimated via learning from information collected during navigation.\footnote{The terms $P_S$, $R_S$, and $R_E$ are implicitly functions of the belief, but shown here only as functions of the chosen action for notational simplicity.}

In the LSP approach~\cite{pmlr-v87-stein18a} and in other LSP-derived planners so far~\cite{NEURIPS2021_926ec030, bradley2021}, learning has relied only on \emph{local} information---e.g., semantic information, images, or local structure.
However, locally-accessible information alone cannot inform effective predictions about unseen space in general; information revealed elsewhere in the environment may determine where a robot should navigate next.
As such, the learned models upon which existing LSP approaches rely perform poorly in even simple environments where non-locally available information is required.
We show one example of this limitation in Sec.~\ref{sec:example-case} and discuss how we use Graph Neural Networks to overcome it in Sec.~\ref{sec:eq-theory-non-local}.
\section{Motivating Example: A Memory Test for Navigation}
\label{sec:example-case}
\begin{figure}[t] \vspace{3pt}
  \includegraphics[width=.48\textwidth]{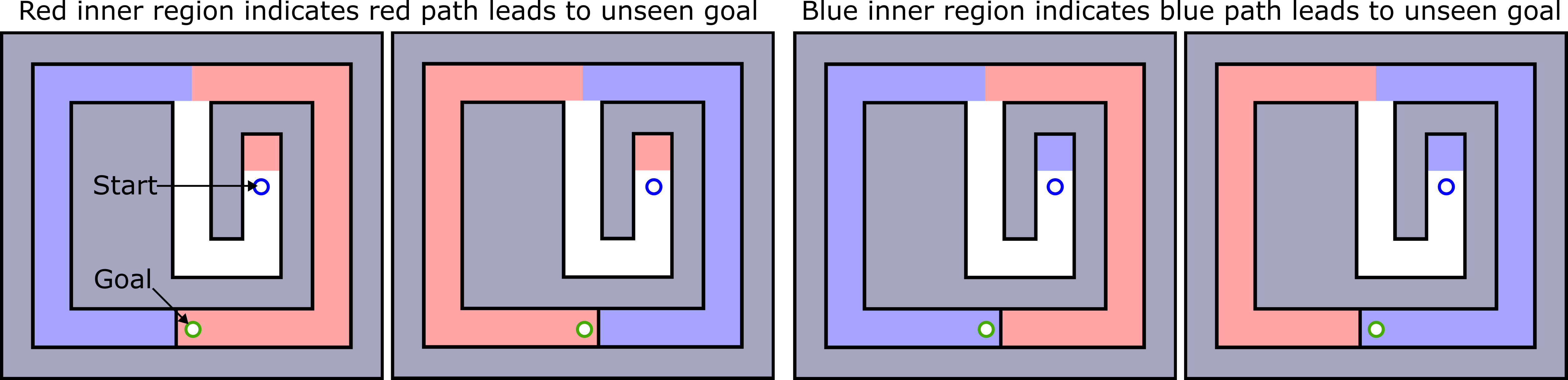}
    \caption{
    \textbf{Low cost navigation in our J-Intersection environment requires non-local information.}
    When the goal is either on left or right from the intersection, we need the non-local information from the start position to decide correctly at the intersection. Choosing always left or right or even choosing one color over another will not reliably succeed.
    }
  \label{fig:example-case-j-shaped}
\end{figure}

Fig.~\ref{fig:example-case-j-shaped} shows an example scenario motivating the necessity of using non-locally observable information to make good predictions about the environment while trying to reach the goal under uncertainty.
Our \emph{J-Intersection} environment has either a red or blue square region inside of it and around the corner occluded from that square region far away at the intersection that colored region leads to the goal (bottom).

Maps in this environment are structured so that the color of the hallway the robot should follow matches the color of the center region of the map.
We randomize the color of the center map region and mirror the environment randomly so that no systematic policy (e.g., \emph{follow the blue hallway} or \emph{turn left at the fork}) will efficiently reach the goal.

Since the LSP approach is limited to making predictions for the subgoal using only locally observable information, it cannot to learn the (simple) defining structural characteristic of the environment: if the inside square region is red then the path to the goal is red and if the inside square is blue then blue is the path to the goal.
Instead, we will augment the LSP approach to rely on a \emph{graph neural network}~\cite{peter2018} to estimate the subgoal properties, allowing it to use both local and non-local information to make predictions about the goodness of actions that enter unseen space and thus perform well across a variety of complex environments.
\section{Approach: Making Predictions about Unseen Space using Non-Local Information}\label{sec:eq-theory-non-local}
We aim to improve navigation under uncertainty by estimating task-relevant properties of unseen space via non-locally observable information.
Consistent with our discussion in Sec.~\ref{sec:lsp} for modelling uncertainty via POMDP, our robot relies on the LSP model-based planning abstraction of Stein et al.~\cite{pmlr-v87-stein18a} for high-level navigation through partially-revealed environments, for which learning is used to estimate the \emph{subgoal properties} ($P_S$, $R_S$, and $R_E$) used to determine expected cost via Eq.~\eqref{eq:lsp-planning}.

We will use a Graph Neural Network (GNN) to overcome the limitations of making predictions using only local information (as discussed in Sec.~\ref{sec:example-case}) and thus improve both predictive power and planning performance.
A graph-based representation of the environment captures both topological structure and also allows information to be retained and communicated over long distances~\cite{fengda2021, azizi2021}.
A GNN is a deep-learning approach that allows predictions over graph data; to plan, we require estimates of the properties ($P_S$, $R_S$, and $R_E$) for each subgoal node and so our graph neural network will output estimates of these properties for each.
In the following sections, we detail how we convert the environment into a graph representation (Sec.~\ref{sec:generate-graph}), how training data is generated (Sec.~\ref{sec:gen-train-data}), and the network and training parameters (Sec.~\ref{sec:gnn-parameters}).

\subsection{Computing a High-level Graph Representation}
\label{sec:generate-graph}
While the occupancy grid of the observed region can be used as a graph representation of the environment, it has too many nodes for learning to be practical.
Instead, we want to generate a simplified (few-node) graph of the environment that preserves high-level topological structure, so that nodes exist at (i) intersections, (ii) dead-ends, and (iii) subgoals.

\begin{figure}[t] \vspace{3pt}
  \includegraphics[width=.48\textwidth]{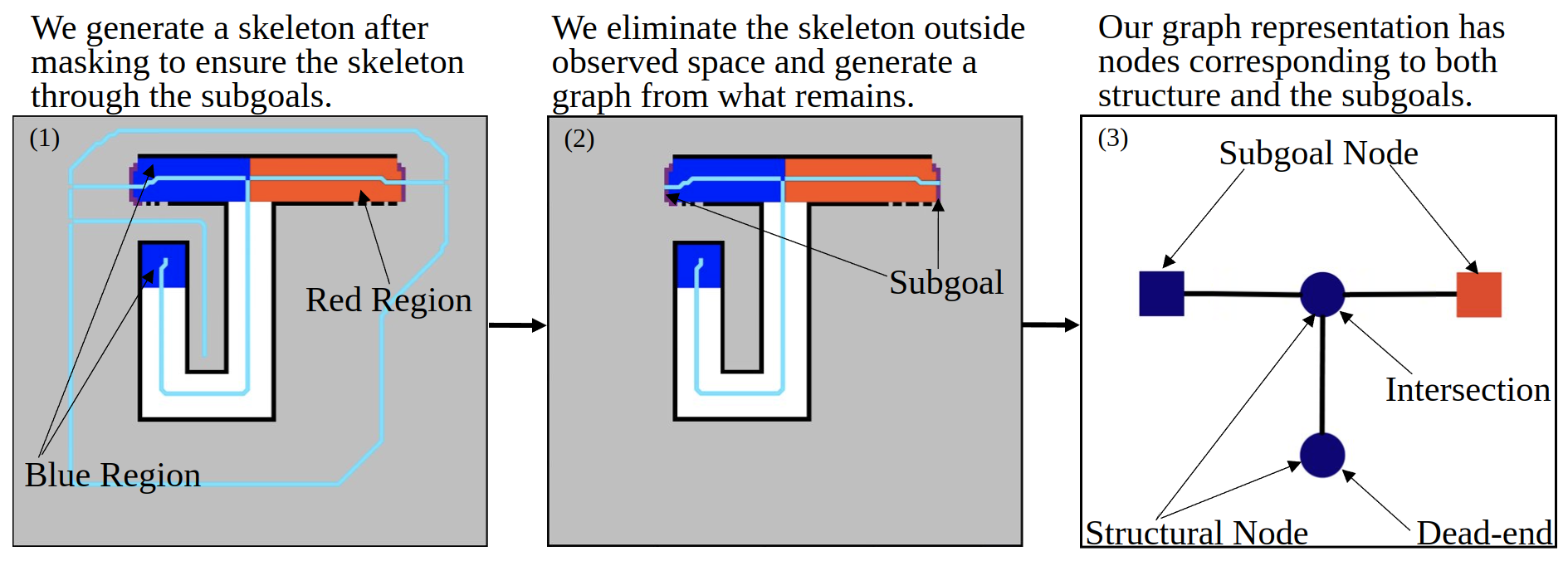}
  \caption{\textbf{Graph representations of the environment for our graph neural net are computed from the partial map.} We use an image skeleton~\cite{zhang1984} to generate a graph from the partial occupancy grid. See Sec.~\ref{sec:generate-graph} for details.
  }
  \label{fig:masked-graph}
\end{figure}

\textit{Graph Generation:}
We create this graph via a process shown in Fig.~\ref{fig:masked-graph}.
We first generate a skeleton~\cite{zhang1984,krishna2016} over a modified version of the map in which unknown space is marked as free yet where frontiers are masked as obstacles except for a single point near their center.
We eliminate the skeleton outside known space and add nodes at all intersections and skeleton endpoints and finally use the skeleton to define the edges between them.
We additionally add nodes corresponding to each subgoal and connect each new node to its nearest structural neighbor in the graph generated from the skeletonization process.
Finally, we add a \emph{goal node} at the location of the goal that has an edge connection to every other node; this \emph{global node}~\cite{peter2018} allows for the propagation of information across the entire environment.

\textit{Neural Network Input Features:}
Structure alone is often insufficient to inform good predictions of unseen space. 
As such, we seek to not only compute a topometric graph of the environment, but also associate semantic information with each node.
Each graph node is given a local observation---a \emph{node feature}---from which the subgoal properties ($P_S$, $R_S$, and $R_E$ in Eq.~\eqref{eq:lsp-planning}) will be estimated via the graph neural network.
Node features are 6-element vectors: (i) a 3-element one-hot semantic class (or color) at the location of the node, (ii) the number of neighbors of that node, (iii) a binary indicator of whether or not the node is a subgoal, and (iv) a binary indictor of whether the node is the goal node.
We additionally include a single edge feature, associated with each edge in the graph: the geodesic distance between the nodes it connects.
Owing to the presence of a goal node connected to every other node, the edge features provides each node its distance to the goal.
To ensure a fair comparison with the LSP-Local planner, our learned baseline that does not consider edge information, the node features for LSP-Local are augmented to include the geodesic distance to the goal.
Conditioned upon, correctly building the map the input is enough to ensure safety during navigation.
Safety during navigation with the aforementioned inputs is ensured conditioned upon correctly building the maps.

\subsection{Graph Neural Network Structure and Training}
\label{sec:gnn-parameters}
We use the PyTorch~\cite{pytorch} neural network framework and Torch Geometric~\cite{torch-geometric} to define and train our graph neural network.
The neural network begins with 3 locally-fully-connected layers, which are fully-connected layers that processes the features for each node in isolation, without considering the edges or passing information to neighbors; all three have hidden layer dimension of 8.
Next, the network has 4 GATv2Conv~\cite{GATv2Conv} layers, each with hidden layer dimension of 8.
Finally, a locally-fully-connected layer takes in the 8-dimensional node features as input and produces a three dimensional output: a logit corresponding to $P_S$ and the two cost terms $R_S$ and $R_E$.
For the LSP-Local learned-baseline planner, we replace the GATv2Conv graph neural network layers with locally-fully-connected layers, eliminating sharing of information between nodes and thus its ability to use non-locally-available information to make predictions about unseen space.


\textit{Loss Function:}
Our loss function matches the original LSP approach of Stein et al.~\cite{pmlr-v87-stein18a} adapted for our graph input data.
For each subgoal node, we accumulate error according to a weighted cross-entropy loss (a classification objective) for $P_S$ and an L1-loss (a regression objective) for $R_S$ and $R_E$.
Since only the properties of the subgoal nodes are needed, we mask the loss for non-subgoal nodes and only consider the subgoal nodes' contribution to the loss.

\textit{Training Parameters:}
We train a separate network (with identical parameters) for each environment. 
Training proceeds for 50k steps.
The learning rate begins at $10^{-3}$ and decays by a factor of 0.6 every 10k steps.

\subsection{Generating Training Data}
\label{sec:gen-train-data}
To train our graph neural network, we require training data collected via offline navigation trials from which we can learn to estimate the subgoal properties ($P_S$, $R_S$, and $R_E$) for each subgoal node in the graph.
During an offline training phase, we conduct trials in which the robot navigates from start to goal and generates labeled data at each time step.
Training data consists of environment graphs $G$---with input features consistent with our discussion in Sec.~\ref{sec:generate-graph}---and labels associated with each subgoal node.

To compute the labels for our training data, we use the underlying known map to determine whether or not a path to the goal exists through a subgoal.
Using this information, we record a label for each subgoal that corresponds to a sample of the probability of success $P_S$ and from which we can learn to estimate $P_S$ using cross-entropy loss.
Labels for the other subgoal properties are computed similarly: labels for the success cost $R_S$ correspond to the travel distance through unknown space to reach the goal, for when the goal can be reached, and the exploration cost $R_E$ is a heuristic cost corresponding to how long it will take a robot to realize a region is a dead end, approximated as the round-trip travel time to reach the farthest reachable point in unseen space beyond the chosen frontier.
This data and collection process mirrors that of LSP~\cite{pmlr-v87-stein18a}; readers are referred to their paper for additional details.

We repeat the data collection process for each step over hundreds of trials for each training environment.
So as to generate more diverse data, we switch between the known-space planner and an optimistic (non-learned) planner to guide navigation during data generation.
The details of each environment can be found in Sec.~\ref{sec:results}.

\section{Experimental Results}
\label{sec:results} 
We conduct simulated experiments in three environments---our \emph{J-Intersection} (Sec.~\ref{sec:results:jint}), \emph{Parallel Hallway} (Sec.~\ref{sec:results:hallways}), and \emph{University Building} (Sec.~\ref{sec:results:office})---in which a robot must navigate to a point goal in unseen space.
For each trial, we evaluate performance of 4 planners:
\begin{LaTeXdescription}
\item[Non-Learned Baseline] Optimistically assumes the unseen space to be free and plans via grid-based A$^{\!*}$ search.
\item[LSP-Local (learned baseline)] Plans via Eq.~\eqref{eq:lsp-planning}, estimating subgoal properties via only local features, as in \cite{pmlr-v87-stein18a}.
\item[LSP-GNN (ours)] Plans via Eq.~\eqref{eq:lsp-planning}, yet uses our graph neural network learning backend to estimate subgoal properties using both local and non-local features.
\item[Fully-Known Planner] The robot uses the fully-known map to navigate; a lower bound on cost.
\end{LaTeXdescription}
For each planner, we compute average navigation cost across many (at least 100) random maps from each environment.

\begin{table}[t]
    \begin{center}
    \caption{Avg. Cost over 100 Trials in the J-Intersection Environment}
    \label{table:toy-stats}
        \begin{tabular}{cc}
        \toprule
            \textbf{Planner} & \textbf{Avg. Cost (grid cell units)} \\
            \hline     
            Non-Learned Baseline & $303.03$\\        
            LSP-Local (learned baseline) & $323.46$\\      
            LSP-GNN (ours) & \textbf{204.85}\\
            \hline
            Fully-Known Planner & $204.85$\\   
        \bottomrule
        \end{tabular}
    \end{center}
\end{table}

\subsection{J-Intersection Environment}\label{sec:results:jint}
\begin{figure}[t]\vspace{3pt}
  \includegraphics[width=.48\textwidth]{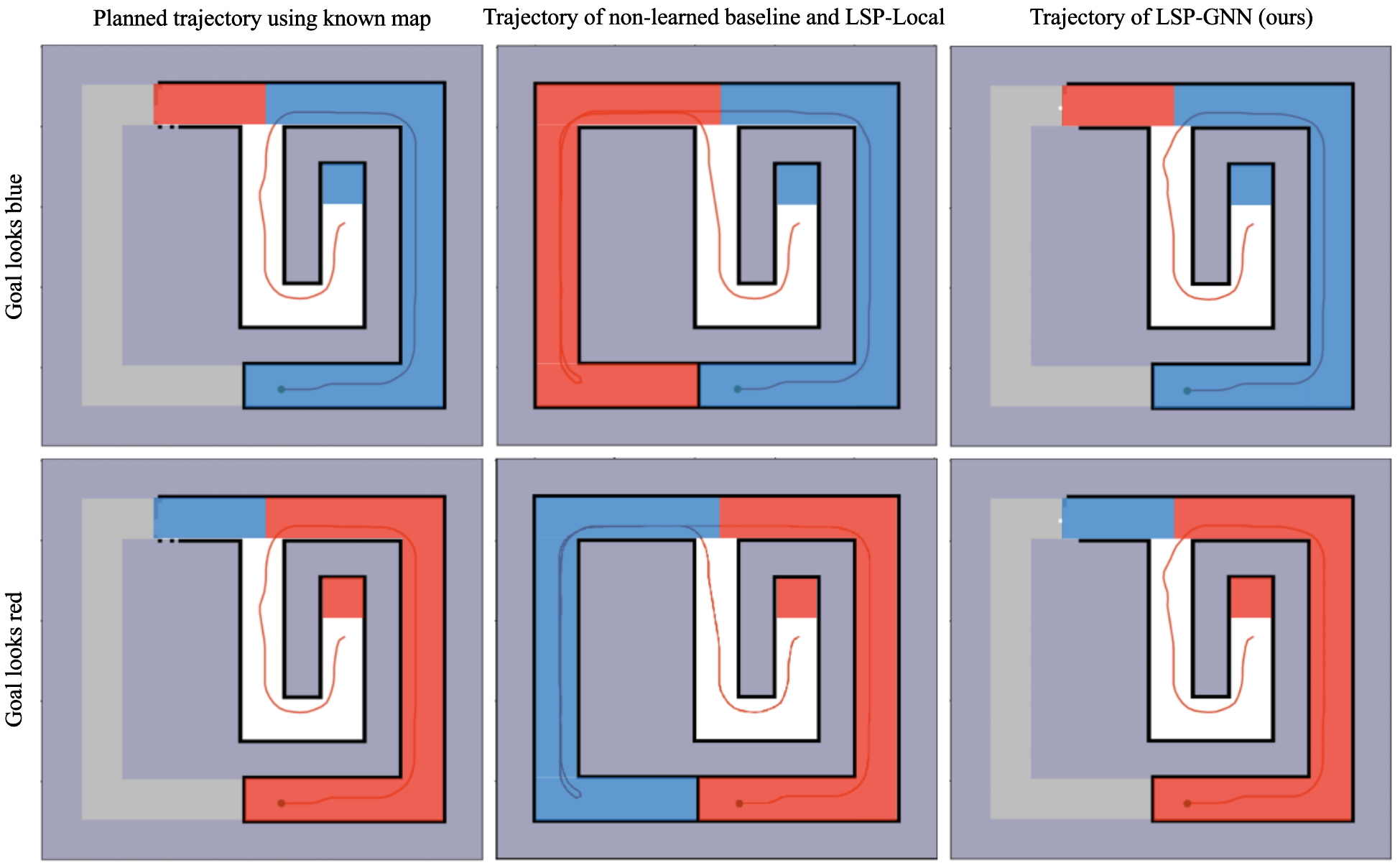}
  \caption{\textbf{Planned trajectories of the bench-marked planner approaches.}
  J-intersection environment where the goal is on the right.
  The left column shows the optimal trajectory (planned using the underlying known map).
  The middle column shows the same trajectory of both the non-learned baseline and LSP-Local where they make a systematic choice.
  The right column shows the trajectory planned by LSP-GNN that is similar to the optimal one.}
  \label{fig:example-case-results}
\end{figure}

We first show results in the J-Intersection environment, described in Sec.~\ref{sec:example-case} to motivate the importance of non-local information for good performance for navigation under uncertainty.
In this environment, the robot must choose where to travel at a fork in the road, yet non-locally observable information is needed to reliably make the correct choice---a blue-colored starting region indicates that the goal can be reached by turning towards the blue hallway at the intersection, and the same for the red-colored regions. We randomly mirror the environment so that the robot cannot learn a systematic policy that quickly reaches the goal without understanding.

We conduct 100 trials for each planner in this environment to evaluate their performance and show the average cost planning strategy in Table~\ref{table:toy-stats}.
Across all trials, our proposed LSP-GNN planner \emph{always} correctly decides where to go at the intersection and achieves near-perfect performance.
By contrast, both the LSP-Local and Non-Learned Baseline planners lack the knowledge to determine which is the correct way to go and perform poorly overall, resulting in poor performance in roughly half of the trials.
We highlight two example trials in Fig.~\ref{fig:example-case-results}.
We do not report the prediction accuracy empirically, because the prediction accuracy does not reflect the actual gain in performance for our work.

\subsection{The Parallel Hallway Environment}\label{sec:results:hallways}
\begin{figure}[t]
  \includegraphics[width=.48\textwidth]{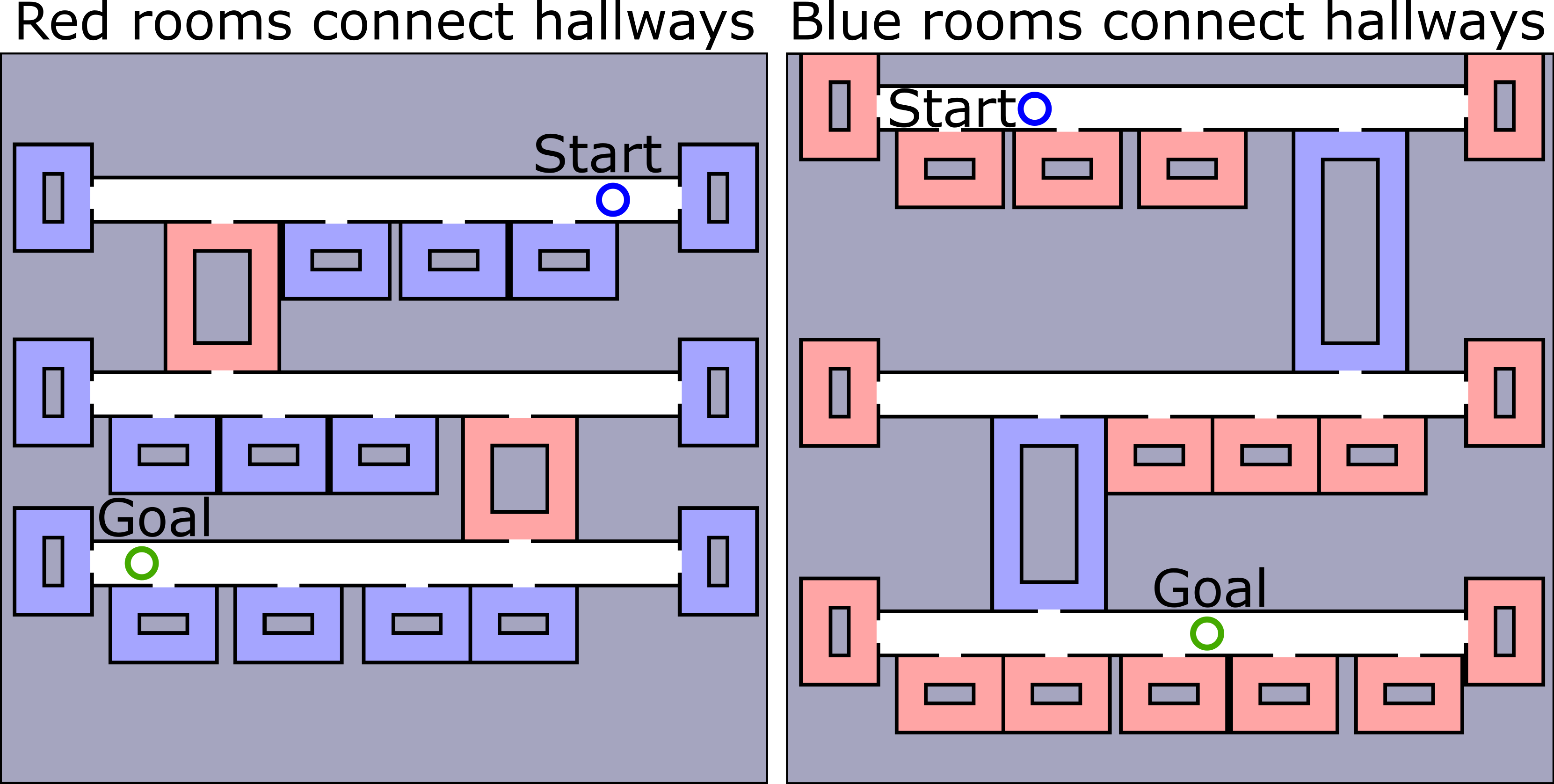}
  \caption{\textbf{Two sample maps from our procedurally-generated Parallel Hallway environment.} A robot is tasked to navigate from start to goal in these maps without having access to the underlying map. The left image shows a sample map where the red rooms connect the hallways and the right image shows where the blue rooms connect the hallways.}
  \label{fig:sample-hallway}
\end{figure}

Our \emph{Parallel Hallway} environment (Fig.~\ref{fig:sample-hallway}) consists of parallel hallways connected by rooms.
We procedurally generate maps in this environment with three hallways and two room types: (i) \emph{dead-end} rooms and (ii) \emph{passage} rooms that provide connections between neighboring parallel hallways.
Only one passage room exists between a pair of hallways, and so the robot must identify this room if it is to travel to another hallway.
Environments are generated such that the dead-end rooms all have the same color (red or blue) distinct from the color of the passage rooms, which are thus blue or red, respectively.
We are making the environment such that the relational information, such as recognizing that if a room with certain color is explored as a dead-end, then the other colored room serves as a pass-through room can be learned. 
If the colors were entirely random, there would be no way to make predictions about the unseen space.
Both room types contain obstructions and are otherwise identical, so that it is not possible to tell whether or not a room will connect to a parallel hallway without trial-and-error or by utilizing semantic color information from elsewhere in the map.
Rooms are placed far enough apart that the robot cannot determine from the local observations if a room will lead to the next hallway or will be a dead end.
The start and goal locations are placed in separate hallways, so as to force the robot to understand its surroundings to reach the goal quickly.
Thus, to navigate well in this challenging procedurally-generated environment, the robot must first explore, trying nearby rooms to determine which color belongs to which room type, and then retain this information to inform navigation through the rest of the environment.

\begin{table}[t]
    \begin{center}  
    \caption{Avg. Cost over 500 Trials in the Parallel Hallway Environment}\label{table:hallway-stats}
        \begin{tabular}{cc}
            \toprule
            \textbf{Planner} & \textbf{Avg. Cost  (grid cell units)}\\
            \hline
            Non-Learned Baseline & $205.93$\\
            LSP-Local (learned baseline) & $236.47$\\
            LSP-GNN (ours) & \textbf{141.37}\\
            \hline
            Fully-Known Planner & $108.37$\\
            \bottomrule
        \end{tabular}
    \end{center} 
\end{table}

\begin{figure}[t]
  \includegraphics[width=.48\textwidth]{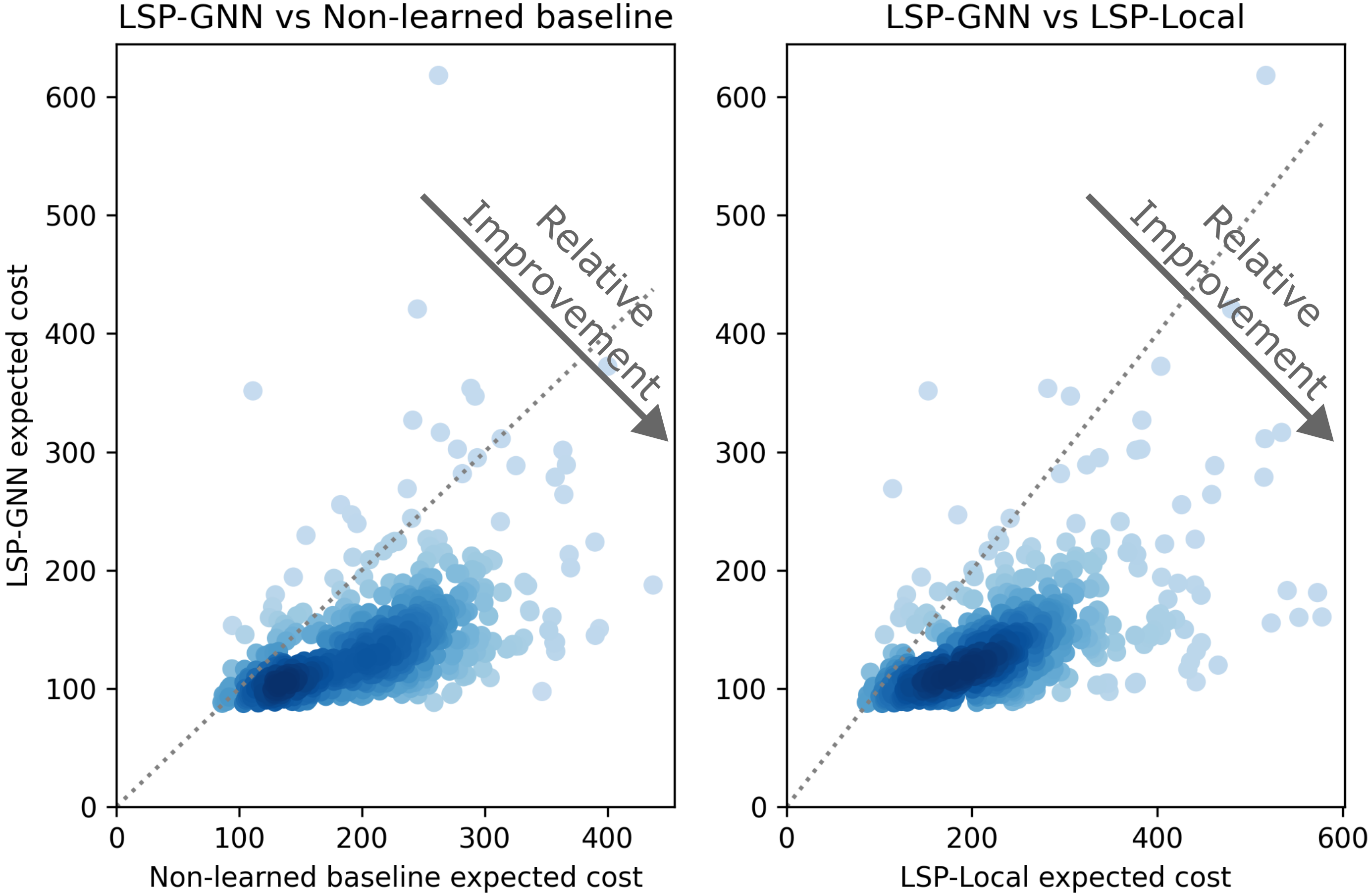}
  \caption{\textbf{Parallel Hallway Results: average cost over 500 trials decreases using LSP-GNN.} Our learning-informed planner outperforms both the non-learned baseline (left) and the LSP-Local (right) planners.}
  \label{fig:scatter-plot-hallway}
\end{figure}

We train the simulated robot on data from 2,000 distinct procedurally generated maps and evaluate in a separate set of 500 distinct procedurally generated maps. We show the average performance of each planning strategy in Table~\ref{table:hallway-stats} and include scatterplots of the relative performance of different planners for each trial in Fig.~\ref{fig:scatter-plot-hallway}.
The robot planning with our LSP-GNN approach is able to utilize non-local local information to improve its predictions about how best to reach the goal, achieving a 31.3\% improvement in average cost versus the optimistic Non-Learned Baseline planner and a 40.2\% improvement over the LSP-Local planner. In addition, our approach is \emph{reliable}: owing to the LSP planning abstraction, our robot is able to successfully reach the goal in all maps.

\begin{figure}[t]
  \includegraphics[width=.48\textwidth]{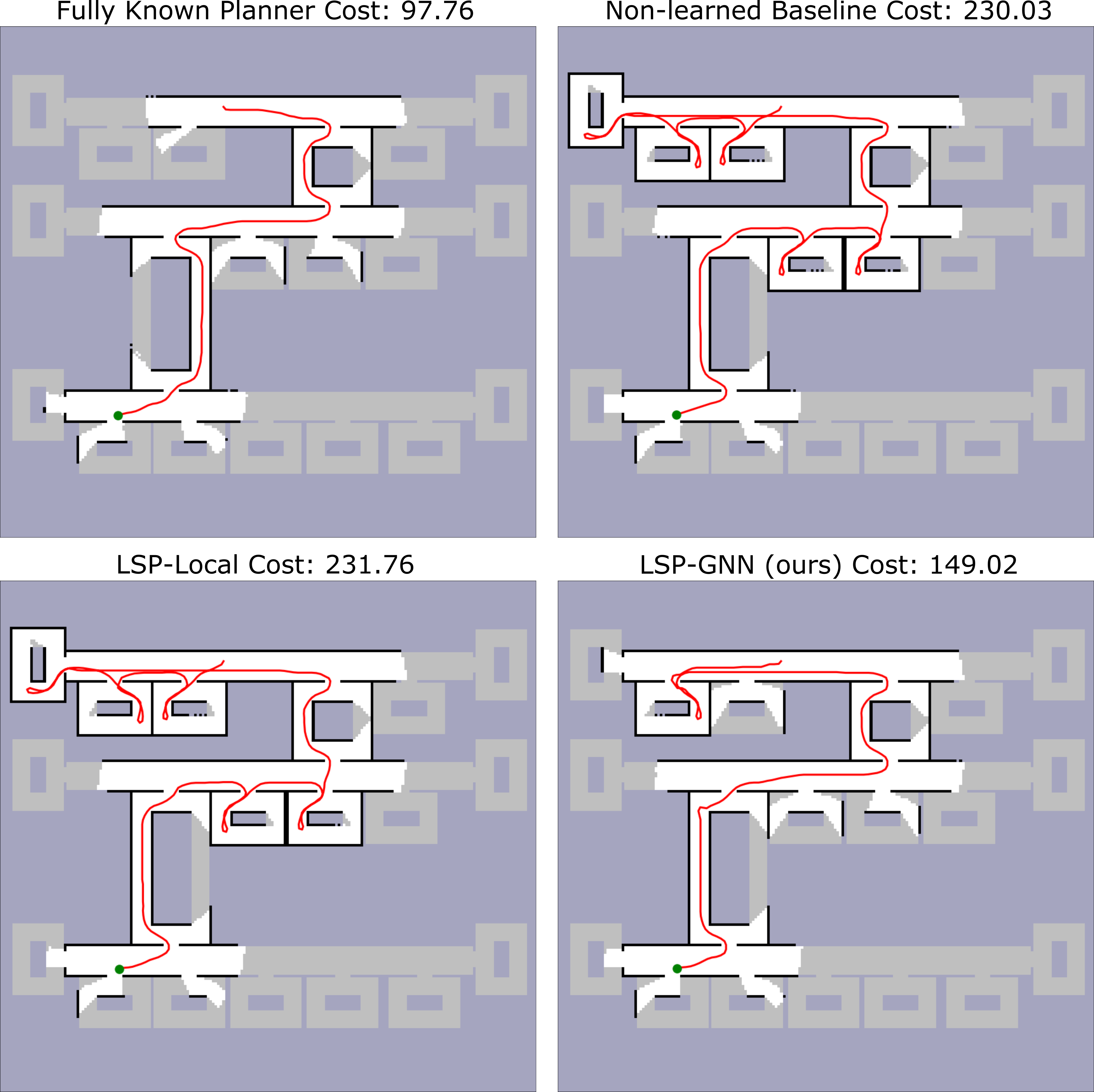}
  \caption{\textbf{Navigation trajectories of the tested planners in one of the testing maps from the parallel hallway environment.}
  Using non-local information enables LSP-GNN to perform better than both the learned (LSP-Local) and non-learned (Dijkstra) baselines.
  }
  \label{fig:result-hall-path}
\end{figure}

We highlight one trial in Fig.~\ref{fig:result-hall-path}, in which the robot is tasked to navigate from the top hallway to the bottom hallway, which contains the goal.
After a brief period of trial-and-error exploration in the first (top) hallway, the robot discovers the passage to the neighboring hall and uses the knowledge of the semantic color to quickly locate the passage to the next hallway and reach the goal.
By contrast, the Non-Learned Baseline optimistically assumes unseen space to be free and enters every room in the direction of the goal.
The LSP-Local planner makes predictions using only local information and, unable to use important navigation-relevant information, cannot determine how to reach the goal; its poor predictions result in frequent turning-back behavior as it seeks alternate routes to the goal, reducing performance.

\subsection{University Building Floorplans}\label{sec:results:office}
\begin{figure}[t]
  \includegraphics[width=.48\textwidth]{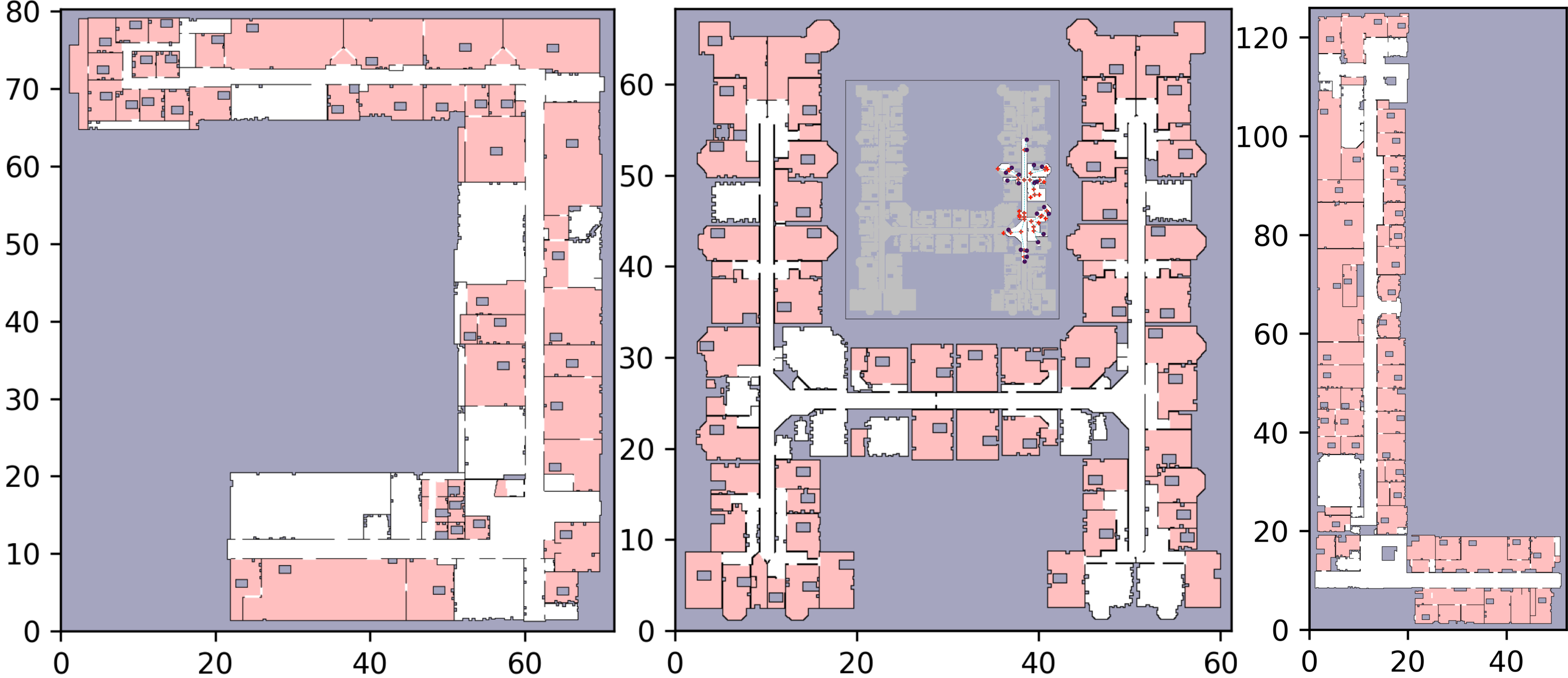}
  \caption{\textbf{Three large-scale training maps from our university floorplan environment, each generated from a real-world floor plan.}
  The inset in the center map shows an instance of a graph (as used to define our graph neural network) for a partial map during a navigation trial. 
  Plot axes are in units of meters.
  }
  \label{fig:mit-train-example-map}
\end{figure}
Finally, we evaluate in large-scale maps generated from real-world floorplans of buildings from the Massachusetts Institute of Technology, including buildings of over 100 meters in extent along either side; see Fig.~\ref{fig:mit-train-example-map} for examples.
We generate data from 2,000 trials across 56 training floorplans and evaluate in 250 trials from 9 held-out test floorplans, each augmented by procedurally generated clutter to add furniture-like obstacles to rooms.
In addition to occupancy information, \emph{rooms} in the map have a distinct semantic class from \emph{hallways} (and other large or accessible spaces); this semantic information is provided as input node features to the neural networks to inform their predictions.

\begin{figure}[t]
  \includegraphics[width=.48\textwidth]{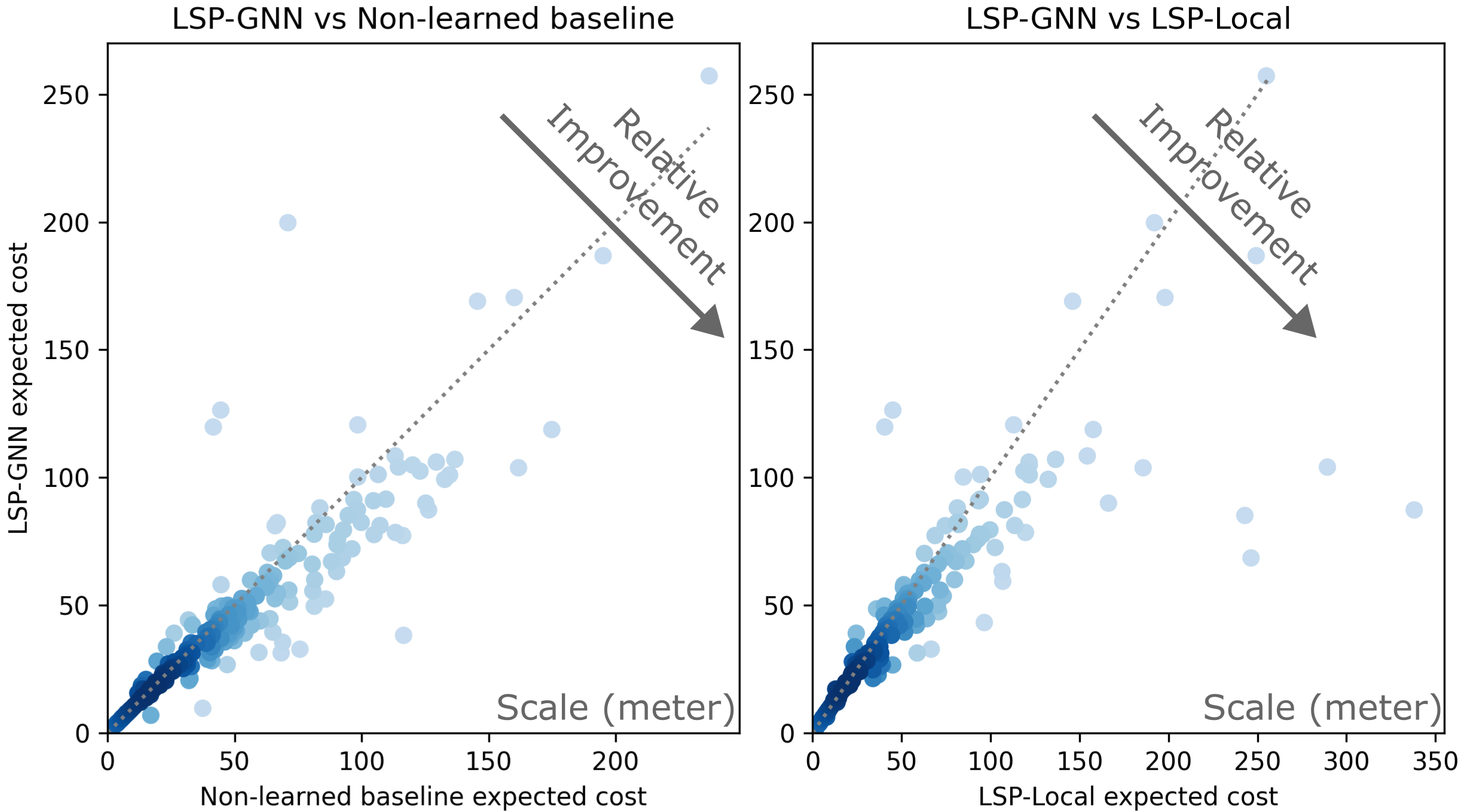}
  \caption{\textbf{University Building Floorplan Results: average cost (meter) over 250 trials decreases using LSP-GNN.} Our learning-informed planner outperforms both the non-learned baseline (left) and the LSP-Local (right) planners.}
  \label{fig:scatter-plot-office}
\end{figure}

We show the average performance of each planning strategy in Table~\ref{table:office-stats} and include scatterplots of the relative performance of different planners for each trial in Fig.~\ref{fig:scatter-plot-office}.
The robot planning with our LSP-GNN approach achieves improvements in average cost of 9.3\% versus the optimistic Non-Learned Baseline planner and of 14.9\% improvement over the LSP-Local Learned Baseline planner. 
Unlike the LSP-Local planner, which does not have enough information to make good predictions about unseen space, our LSP-GNN approach can make use of non-local information to inform its predictions and thus performs well despite the complexity inherent in these large-scale testing environments. 


\begin{table}[t]
    \begin{center}  
    \caption{Avg. Cost over 250 Trials in the University Building Floorplans}\label{table:office-stats}
        \begin{tabular}{cc}
            \toprule
            \textbf{Planner} & \textbf{Avg. Cost (meter)}\\
            \hline
            Non-Learned Baseline & $44.98$\\
            LSP-Local (learned baseline) & $47.93$\\
            LSP-GNN (ours) & \textbf{40.80}\\
            \hline
            Fully-Known Planner & $31.77$\\
            \bottomrule
        \end{tabular}
    \end{center} 
\end{table}

\begin{figure}[t]
  \includegraphics[width=.48\textwidth]{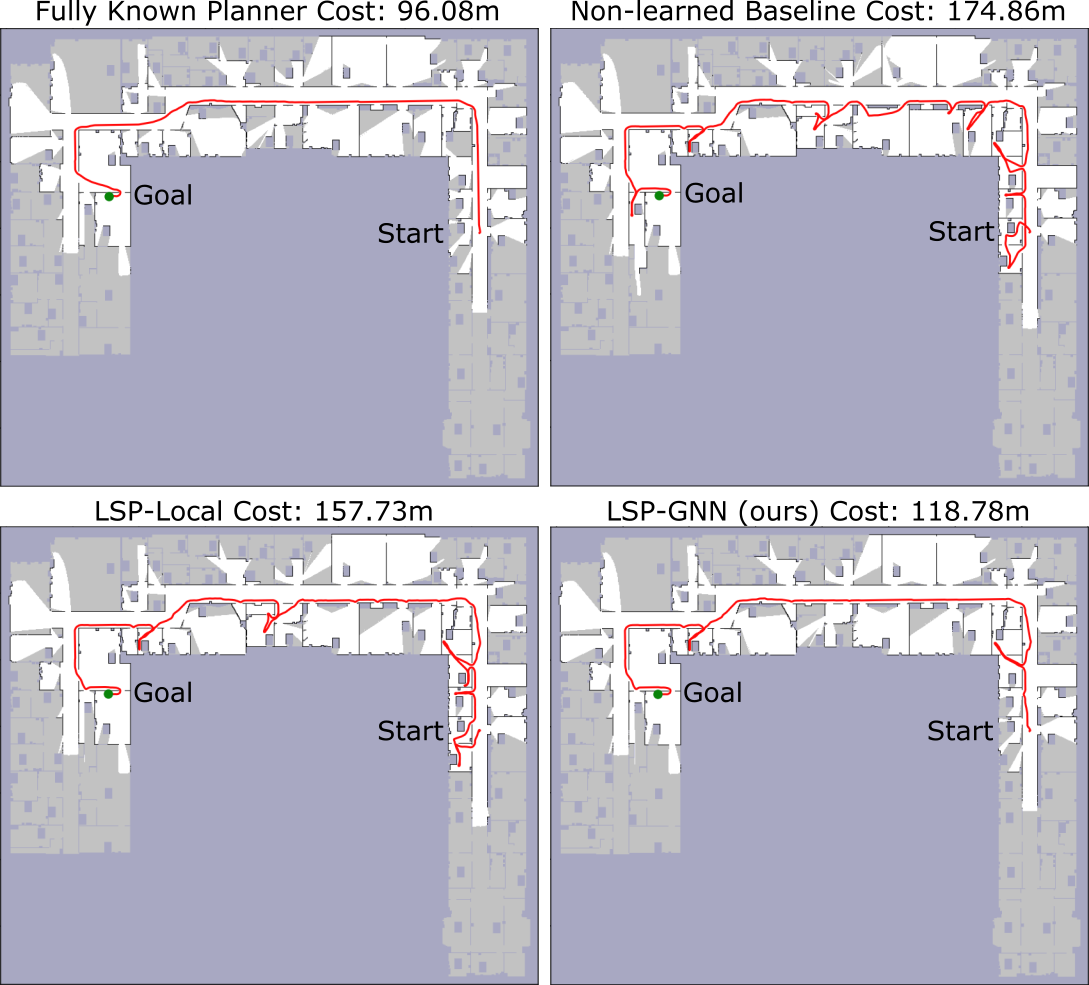}
  \caption{\textbf{Navigation trajectories of all tested planners in one of the large-scale testing maps from the university building environment.}
  LSP-GNN performs better than both the learned (LSP-Local) and non-learned (Dijkstra) baselines deviating very few times from the hallway to reach faraway goal.
  }
  \label{fig:result-office-good}
\end{figure}

\begin{figure}[t]
  \includegraphics[width=.48\textwidth]{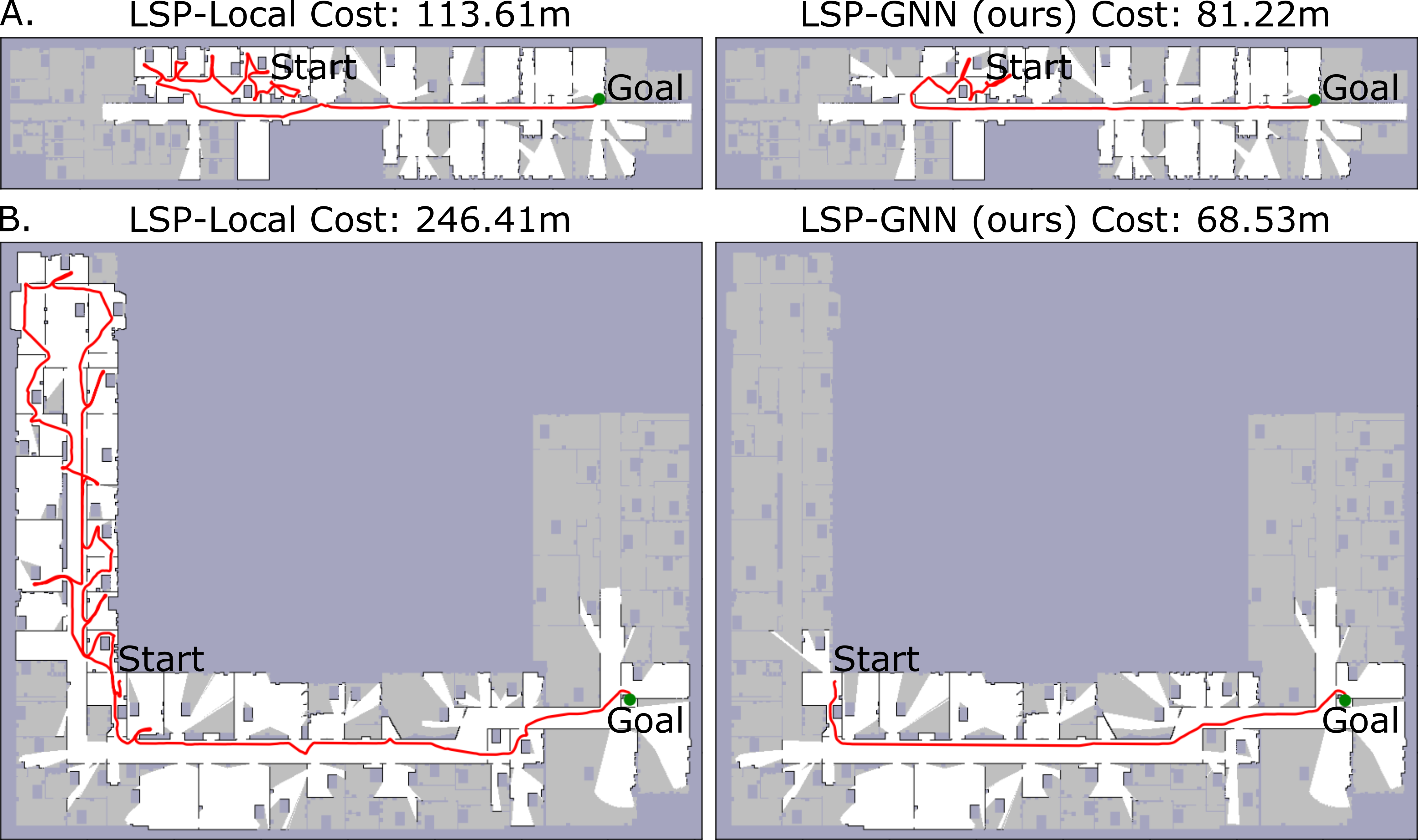}
  \caption{
  \textbf{Two comparison between the navigational trajectories of our LSP-GNN against LSP-Local.}
  LSP-GNN exhibits the capacity to recover quickly than LSP-Local when both planner cannot immediately find the correct path. 
  }
  \label{fig:result-office-bad}
\end{figure}

Fig.~\ref{fig:result-office-good} shows a typical navigation example in one of our test environments. 
In this scenario, the shortest possible trajectory involves knowing to follow hallways until near to the goal.
Both learned planners generally exhibit hallway-following behavior---often useful in building-like environments such as these---and improve upon the non-learned (optimistic) baseline. However, our LSP-GNN planner, able to make use of non-local information, can more reliably determine which is the more productive route and more quickly reaches the faraway goal.
Fig.~\ref{fig:result-office-bad} shows two additional examples that highlight the improvements of our LSP-GNN planner made possible by non-locally-available information. In Fig.~\ref{fig:result-office-bad}A, we highlight an example in which both learned planners cannot immediately find the correct path, yet LSP-GNN is able to improve its predictions about where is most likely to lead to the unseen goal and recover more quickly than does LSP-Local. Fig.~\ref{fig:result-office-bad}B shows a more extreme example, in which the LSP-Local planner fails to quickly turn back to seek a promising alternate route immediately identified by LSP-GNN.

\section{Conclusion and Future Work}\label{sec:conclusion-future}
We present a reliable model-based planning approach that uses a graph neural network to estimate the goodness of goal-directed high-level actions from both local and non-local information, improving navigation under uncertainty.
Our planning approach takes advantage of non-local information to make informed decisions about how to more quickly reach the goal.
We rely on a graph neural network (GNN) to make these predictions. The GNN consumes a graph representation of the partial map and makes predictions about the goodness of potential routes to the goal.
We demonstrate improved performance on two simulated environments in which non-local information is required to plan well, demonstrating the efficacy of our approach.

In future work, we envision passing more complex sensory input to the robot, allowing it to estimate the goodness of its actions using information collected from image sensors or semantically-segmented images.

\bibliographystyle{IEEEtran}
\bibliography{main}

\end{document}